\documentclass[11pt,a4paper]{article}
\usepackage[hyperref]{acl2020}
\usepackage{times}
\usepackage{latexsym}
\usepackage{multirow}

\usepackage{microtype}

\aclfinalcopy

\usepackage{amsmath}
\usepackage{amssymb}
\usepackage{graphicx}

\newcommand*\samethanks[1][\value{footnote}]{\footnotemark[#1]}

\title{Re-translation versus Streaming for Simultaneous Translation}

\author{Naveen Arivazhagan\thanks{~~Equal contributions}, Colin Cherry\samethanks, Wolfgang Macherey and George Foster \\
Google Research\\
  {\tt \{navari,colincherry,wmach,fosterg\}@google.com} \\}

\date{}

\newcommand\bias[0]{\beta}
\newcommand\EDAL[0]{DAL}

\begin{document}
\maketitle
\begin{abstract}

There has been great progress in improving streaming machine translation, a simultaneous paradigm where the system appends to a growing hypothesis as more source content becomes available. 
We study a related problem in which revisions to the hypothesis beyond strictly appending words are permitted.
This is suitable for applications such as live captioning an audio feed.
In this setting, we compare custom streaming approaches to re-translation, a straightforward strategy where each new source token triggers a distinct translation from scratch.
We find re-translation to be as good or better than state-of-the-art streaming systems, even when
operating under constraints that allow very few revisions.
We attribute much of this success to a previously proposed data-augmentation technique that adds prefix-pairs to the training data,
which alongside wait-$k$ inference forms a strong baseline for streaming translation.
We also highlight re-translation's ability to wrap arbitrarily powerful MT systems with an experiment showing large improvements from an upgrade to its base model.
\end{abstract}

\section{Introduction}

In simultaneous machine translation, the goal is to translate an incoming stream of source words with as low latency as possible. A typical application is speech translation, where we often assume the eventual output modality to also be speech. In a speech-to-speech scenario, target words must be appended to existing output with no possibility for revision. The corresponding translation task, which we refer to as {\em streaming translation}, has received considerable recent attention, generating custom approaches designed to maximize quality and minimize latency \cite{Cho16,Gu2017,Dalvi2018,ma-etal-2019-stacl}.  However, for applications where the output modality is text, such as live captioning, the prohibition against revising output is overly stringent.

The ability to revise previous partial translations makes simply re-translating each successive source prefix a viable strategy. Compared to streaming models, re-translation has the advantage of low latency, since it always attempts a translation of the complete source prefix, and high final-translation quality, since it is not restricted to preserving previous output. It has the disadvantages of higher computational cost, and a high revision rate, visible as textual \textit{instability} in an online translation display. When revisions are an option, it is unclear whether one should prefer a specialized streaming model or a re-translation strategy.

In light of this, we make the following contributions: (1) We evaluate a combination of re-translation techniques that have not previously been studied together. (2) We provide the first empirical comparison of re-translation and streaming models, demonstrating that re-translation operating in a very low-revision regime can match or beat the quality-latency trade-offs of streaming models. (3) We test a 0-revision configuration of re-translation, and show that it is surprisingly competitive, due to the effectiveness of data augmentation with prefix pairs. 

\section{Related Work}
\label{sec:relwork:stream}

\newcite{Cho16} propose the first streaming techniques for NMT, using heuristic agents based on model scores,
while \newcite{Gu2017} extend their work with agents learned using reinforcement learning.
\newcite{ma-etal-2019-stacl} recently broke new ground by integrating
their read-write agent directly into NMT training.
Similar to \newcite{Dalvi2018}, they employ a simple agent that first reads $k$ source tokens, and then proceeds to alternate between writes and reads until the source sentence has finished.
This agent is easily integrated into NMT training, which allows the NMT engine
to learn to anticipate occasionally-missing source context.
We employ their wait-$k$ training as a baseline, and use their wait-$k$ inference to improve re-translation.
Our second and strongest streaming baseline is the MILk approach of \newcite{arivazhagan-etal-2019-monotonic}, who improve upon wait-$k$ training with an attention that can adapt how it will wait based on the current context. Both wait-$k$ training and MILk attention provide hyper-parameters to control their quality-latency trade-offs: $k$ for wait-$k$, and latency weight
for MILk.

Re-translation was originally investigated by \newcite{niehues2016dynamic,Niehues2018}, and more recently extended by \newcite{arivazhagan2019retranslation}, who propose a suitable evaluation framework, and use it to assess inference-time re-translation strategies for speech translation. We adopt their inference-time heuristics to stabilize re-translation, and extend them with prefix training from \newcite{Niehues2018}. 
Where they experiment on TED talks, compare only to vanilla re-translation and use proprietary NMT, we follow recent work on streaming by using WMT training and test data, and provide a novel comparison to streaming approaches.

\section{Metrics}
\label{sec:metrics}
\begin{table*}[t]
\begin{center}
\begin{tabular}{|l|cccccccc|c|}
\hline
Source & \multicolumn{8}{|l|}{Output} & Erasure \\
\hline
1: Neue & New &&&&&&&& - \\
2: Arzneimittel & New & Medicines &&&&&&& 0 \\
3: k\"onnten & New & Medicines &&&&&&& 0 \\
4: Lungen- & New & drugs & may & be & lung &&&& 1  \\
5: und & New & drugs & could & be & lung & and &&& 3 \\
6: Eierstockkrebs & New & drugs & may & be & lung & and & ovarian & cancer & 4 \\
7: verlangsamen & New & drugs & may & slow & lung & and & ovarian & cancer & 5 \\
\hline
Content Delay & 1 & 4 & 6 & 7 & 7 & 7 & 7 & 7 & \\
\hline
\end{tabular}
\end{center}
\caption{An example prefix translation list for the tokenized German sentence, 
``Neue Arzneimittel k\"onnten Lungen- und Eierstockkrebs verlangsamen'',
with reference, ``New drugs may slow lung , ovarian cancer''. 
}
\label{tab:prefix_list}
\end{table*}

We adapt the evaluation framework from \newcite{arivazhagan2019retranslation}, which includes metrics for latency, stability, and quality. 
Where they measure latency with a temporal lag, we adopt an established token lag that does not rely on machine speed.

Our evaluation is built around a \textit{prefix translation list} (PTL),
which can be generated for any streaming or re-translation system.
For each token in the source sentence (after merging subwords), this list
stores the tokenized system output.
Table~\ref{tab:prefix_list} shows an example.
We use $I$ for the final number of source tokens,
and $J$ for the final number of target tokens.

\subsection{Quality}
Translation quality is measured by calculating BLEU~\cite{Papineni2002} on
the final output of each PTL; that is, standard corpus-level BLEU on
complete translations.
Specifically, we report tokenized, cased BLEU calculated by an internal tool.
We make no attempt to directly measure the quality of intermediate outputs;
instead, their quality is captured indirectly through final output quality and stability.

\subsection{Latency}
Latency is the amount of time the target listener spends waiting for their translation.
Most latency metrics are based on a delay vector $g$, where $g_j$ reports how many source tokens were read before writing the $j^\mathit{th}$ target token~\cite{Cho16}.
This delay is trivial to determine for streaming systems,
but to address the scenario where target content can change,
we introduce the notion of \emph{content delay}, which is
closely related to the finalization event index used to calculate time delay in \newcite{arivazhagan2019retranslation}.

We take the pessimistic view that content in flux is useless;
for example, in Table~\ref{tab:prefix_list}, the 4$^{\textit{th}}$ target token first appears in step 4, but only becomes useful in step 7, when it shifts from \textit{be} to \textit{slow}.
Therefore, we calculate delay with respect to when a token \textit{finalizes}.
Let $o_{i,j}$ be the $j^{\mathit{th}}$ token of the $i^{\mathit{th}}$ output
in a PTL; $1 \leq i \leq I$ and $1 \leq j \leq J$.
For each position $j$ in the final output, we define $g_j$ as: 
\begin{equation*}
g_j =  \min_i \textrm{ s.t. } {o}_{i', j'} = {o}_{I, j'} \; \forall i' \geq i \; \mathrm{and} \;  \forall j' \leq j
\end{equation*}
that is, the number of source tokens read before the prefix ending in $j$ took on its final value.
The \textit{Content Delay} row in Table~\ref{tab:prefix_list} shows delays for our running
example.
Note that content delay is identical to standard delay for streaming systems,
which always have stable prefixes.

With this refined $g$, we can make several latency metrics content-aware, including average proportion~\cite{Cho16},
consecutive wait~\cite{Gu2017}, average lagging~\cite{ma-etal-2019-stacl},
and differentiable average lagging~\cite{arivazhagan-etal-2019-monotonic}.
We opt for differentiable average lagging (DAL) because of its interpretability
and because it sidesteps some problems with average lagging~\cite{Cherry2019}.
It can be thought of as the average number of source tokens a system lags
behind a perfectly simultaneous translator:
\begin{equation*}
  \label{eq:AL}
  \mathrm{\EDAL} = \frac{1}{J}\sum_{j=1}^{J} \left[g'_j-\frac{j-1}{\gamma}\right]
\end{equation*}
where $\gamma = J/I$ accounts for the source and
target having different lengths, and $g'$ adjusts $g$
to incorporate a minimal time cost of $\frac{1}{\gamma}$ for each token:
\begin{equation*}
  g'_j = \left\{
  \begin{array}{ll}
    g_j & j=1\\
    \max\big[g_j, g'_{j-1}+\frac{1}{\gamma}\big] & j>1
  \end{array}\right.
\end{equation*}
Note that DAL sums over the final number of target tokens ($J$),
but it is possible for intermediate hypotheses to have more than $J$ tokens.
Any such tokens are ignored by DAL.

\subsection{Stability}
\label{sec:erasure}
Following \newcite{niehues2016dynamic, Niehues2018} and \newcite{arivazhagan2019retranslation}, we measure stability with \textit{erasure}, which
measures the length of the suffix that is deleted to produce the next revision.
Let $o_i$ be the $i^{\mathit{th}}$ output of a PTL. The normalized erasure (NE) for PTL is defined as:
\begin{equation*}
\mathrm{NE} = \frac{1}{J} \sum_{i=2}^I \big[|o_{i-1}| - |\mathrm{LCP}(o_i, o_{i-1})|\big]
\end{equation*}
where the $|\cdot|$ operator returns the length of a token sequence, and $\mathrm{LCP}$
calculates the longest common prefix of two sequences.
Table~\ref{tab:prefix_list} shows pointwise erasures for each output; its NE would be $13/8=1.625$,
interpretable as the number of intermediate tokens deleted for each final token.

\section{Re-translation Methods}

To evaluate re-translation, we build up the source sentence one token at a time,
translating each resulting source prefix from scratch to construct the PTL for evaluation.

\subsection{Prefix Training}\label{sec:PropAlign}

Standard models trained on full sentences are unlikely to perform well when applied to prefixes. We alleviate this problem by generating prefix pairs from our parallel training corpus, and subsequently training on a 1:1 mix of full-sentence and prefix pairs~\cite{Niehues2018,Dalvi2018}.
Following \newcite{Niehues2018}, we augment our training data with prefix pairs created by selecting a source prefix length uniformly at random, then selecting a target length either proportionally according to sentence length, or based on self-contained word alignments.
For the latter, for each source prefix, we attempt to find a target prefix such that all tokens in the source prefix align only to words in the target prefix and vice versa.
In preliminary experiments, we confirmed a finding by \newcite{Niehues2018} that word-alignment-based prefix selection is no better than proportional selection, so we report results only for the proportional method.\footnote{Word-alignment-based prefix selection may become more important when working on more distant language pairs such as English-Japanese.}
An example of proportional prefix training is given in Table~\ref{tab:prop_train}.
With prefix training, we expect intermediate translations of source prefixes to be shorter, and to look more like partial target prefixes than complete target sentences~\cite{Niehues2018}.

\begin{table*}[t]
    \centering
    \begin{tabular}{|ccp{0.75\textwidth}|}
    \hline
        \multirow{2}{*}{Full} 
        & Source &  Die F\"uhrungskr\"afte der Republikaner rechtfertigen ihre Politik mit der Notwendigkeit , den Wahlbetrug zu bek\"ampfen~~[\textit{15 tokens}]\\
        & Target & Republican leaders justified their policy by the need to combat electoral fraud~~[\textit{12 tokens}]\\
        &&\\
        \multirow{2}{*}{Prefix} 
        & Source & Die F\"uhrungskr\"afte der Republikaner rechtfertigen~~[\textit{5 tokens}]\\
        & Target & Republican leaders justified their~~[\textit{4 tokens}]\\
        \hline
    \end{tabular}
    \caption{An example of proportional prefix training. Each example in the minibatch has a 50\% chance to be truncated, in which case, we truncate its source and target to a randomly-selected fraction of their original lengths, 1/3 in this example. No effort is made to ensure that the two halves of the prefix pair are semantically equivalent.}
    \label{tab:prop_train}
\end{table*}

\subsection{Inference-time Heuristics} \label{sec:bias}

To improve stability, \newcite{arivazhagan2019retranslation} propose a combination of biased search and delayed predictions.
Biased search encourages the system to respect its previous predictions 
by modifying search to interpolate between the distribution from the NMT model (with weight $1-\beta$) and the one-hot distribution formed by the system's translation of the previous prefix (with weight $\beta$).
We only bias a hypothesis for as long as it strictly follows the previous translation. No bias is applied after the first point of divergence.

\label{sec:waitk} 
To delay predictions until more source context is available, we adopt \newcite{ma-etal-2019-stacl}'s wait-$k$ approach at inference time.
We implement this by truncating the target to $\max(i-k, 0)$ tokens, where $i$ is the current source prefix length and $k$ is a constant inference-time hyper-parameter.
To avoid confusion with \newcite{ma-etal-2019-stacl}'s \textit{wait-$k$ training}, we refer to wait-$k$ used for re-translation as \textit{wait-$k$ inference}.\footnote{When wait-$k$ truncation is combined with beam search, its behavior is similar to that of \newcite{zheng-etal-2019-speculative}: sequences are scored accounting for ``future'' tokens that will not be shown to the user.}

\section{Experiments}
We use standard WMT14 English-to-French (EnFr; 36.3M sentences) and WMT15 German-to-English (DeEn; 4.5M sentences) data. For EnFr, we use newstest 2012+2013 for development, and newstest 2014 for test. For DeEn, we validate on newstest 2013 and report results on newstest 2015. 
We use BPE~\cite{Sennrich2016} on the training data to construct a 32K-type vocabulary that is shared between the source and target languages. 

\subsection{Models}  
Our streaming and re-translation models are implemented in Lingvo~\cite{Shen2019}, sharing architecture and hyper-parameters wherever possible. Our RNMT+ architecture \cite{Chen2018} consists of a 6 layer LSTM encoder and an 8 layer LSTM decoder with additive attention~\cite{bahdanau2014neural}. Both encoder and decoder LSTMs have 512 hidden units, apply per-gate layer normalization~\cite{Ba2016}, and use residual skip connections after the second layer. 

The models are regularized using a dropout of 0.2 and label smoothing of 0.1~\cite{Szegedy2016}. Models are optimized using 32-way data parallelism with Google Cloud's TPUv3, using Adam~\cite{Kingma2015} with the learning rate schedule described in \newcite{Chen2018} and a batch size of 4,096 sentence-pairs. Checkpoints for the base models are selected based on development perplexity.

\paragraph{Streaming} We train several wait-$k$ training and MILk models to obtain a range of quality-latency trade-offs. Five wait-$k$ training models are trained with sub-word level waits of 2, 4, 6, 8, and 10. Five MILk models are trained with latency weights of 0.1, 0.2, 0.3, 0.4, 0.5 and 0.75; weights lower than 0.1 tend to increase lag without improving BLEU.
All streaming models use unidirectional encoders and greedy search.

\paragraph{Re-translation}
We test two NMT architectures with re-translation: a
\textit{Base} system
with unidirectional encoding and greedy search, designed for fair comparisons to our streaming baselines above; 
and a more powerful \textit{Bidi+Beam} system using bidirectional encoding and beam search of size 20,
designed to test the impact of an improved base model.
Training data is augmented through the proportional prefix training method unless stated otherwise
(\S~\ref{sec:PropAlign}). Beam-search bias $\bias$ is varied in the range 0.0 to 1.0 in increments of 0.2. When wait-$k$ inference is enabled, $k$ is varied in 1, 2, 4, 6, 8, 10, 15, 20, 30.
Note that we do not need to re-train to test different values of $\beta$ or $k$.

\subsection{Translation with few revisions}\label{qsvs}

\label{sec:res:stream}
\begin{figure*}[ht]
    \centering
    \includegraphics[width=0.47\textwidth]{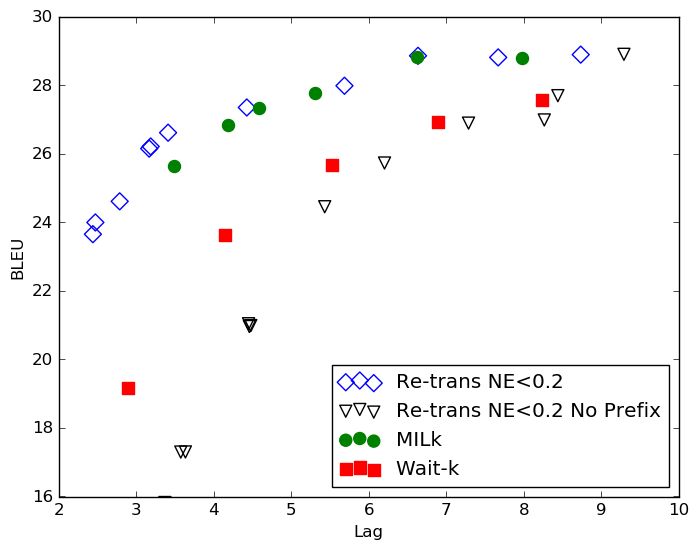}
    \hspace{2em}
    \includegraphics[width=0.47\textwidth]{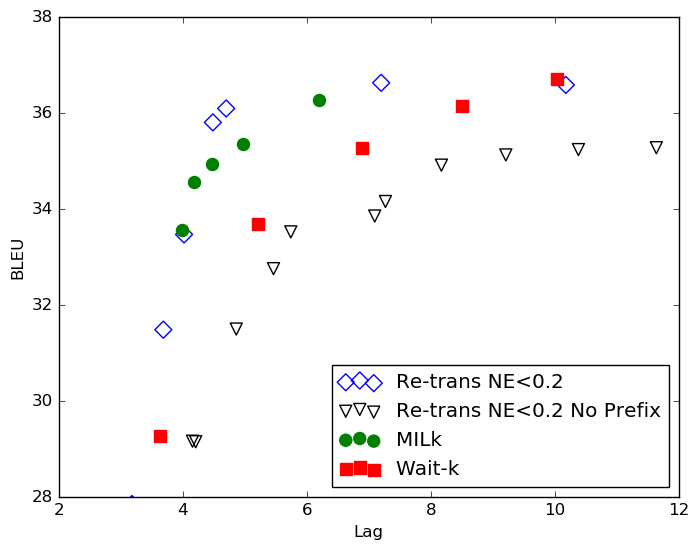}
    \caption{BLEU vs lag (\EDAL{}) curves for translation with low erasure for DeEn (left) and EnFr (right) test sets. 
    }
    \label{fig:bleu_vs_eal}
\end{figure*}
\begin{figure*}[ht]
    \centering
    \includegraphics[width=0.47\textwidth]{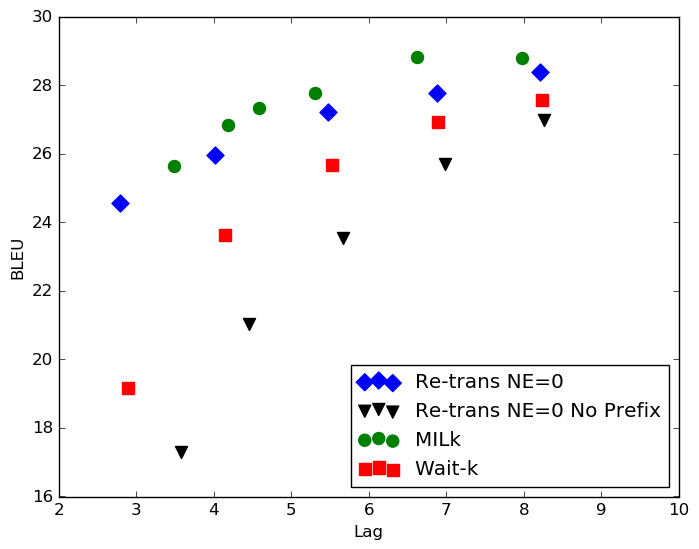}
    \hspace{2em}
    \includegraphics[width=0.47\textwidth]{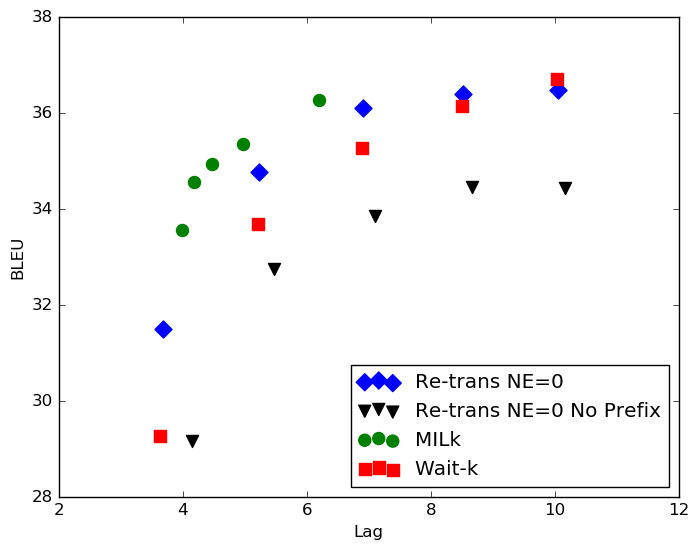}
    \caption{BLEU vs lag (DAL) curves for translation with no erasure for DeEn (left) and EnFr (right) test sets.}
    \label{fig:bleu_lag_sim_streaming}
\end{figure*}

Biased search and wait-$k$ inference used together can reduce re-translation's revisions, as measured by normalized erasure (NE in \S~\ref{sec:erasure}), to negligible levels~\cite{arivazhagan2019retranslation}. But how does re-translation compare to competing approaches? To answer this, we compare the quality-latency trade-offs achieved by re-translation in a low-revision regime to those of our streaming baselines.

First, we need a clear definition of low-revision re-translation.
By manual inspection on the DeEn development set, we observe that systems with an NE of 0.2 or lower display many different latency-quality trade-offs. 
But is NE stable across evaluation sets?
When we compare development and test NE for all 50 non-zero-erasure combinations of $\beta$ and $k$,
the average absolute difference is 0.005 for DeEn, and 0.004 for EnFr, indicating that development NE is very predictive of test NE.
This gives us an operational definition of low-revision re-translation as any configuration with a dev NE $<$ 0.2, allowing on average less than 1 token to be revised for every 5 tokens in the system output.

Since we need to vary both $\beta$ and $k$ for our re-translation systems, we plot BLEU versus \EDAL{} curves by finding the Pareto frontier on the dev set, and then projecting to the test set.
To ensure a fair comparison to our baselines, we test only the \textit{Base} system here.
As an ablation, we include a variant that does not use proportional prefixes, and instead trains only on full sentences.

Figure \ref{fig:bleu_vs_eal} shows our results. Re-translation is nicely separated from wait-$k$, and intertwined with the adaptive MILk. In fact, it is noticeably better than MILk at several latency levels for EnFr. Since re-translation is not adaptive, this indicates that being able to make a small number of revisions is quite advantageous for finding good quality-latency trade-offs. On the other hand, the ablation curve, ``Re-trans NE $<$ 0.2 No Prefix'' is much worse, indicating that proportional prefix training is very valuable in this setting. We probe its value further in the next experiment.

\subsection{Translation with no revisions}
\begin{figure*}[ht]
    \centering
    \includegraphics[width=0.47\textwidth]{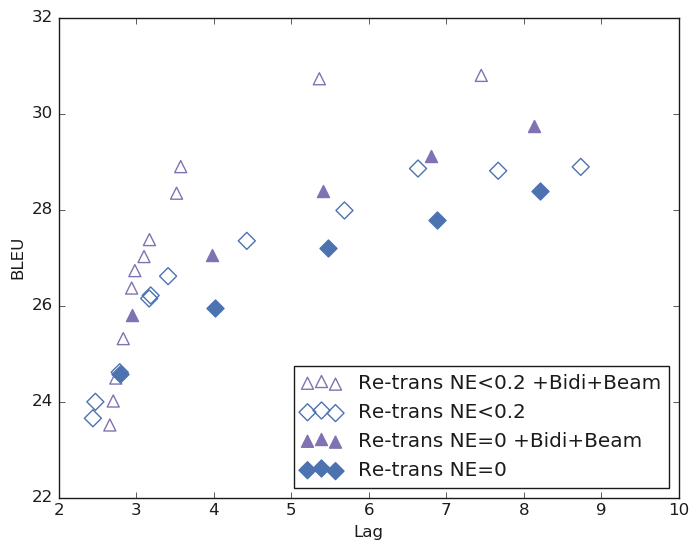}
    \hspace{2em}
    \includegraphics[width=0.47\textwidth]{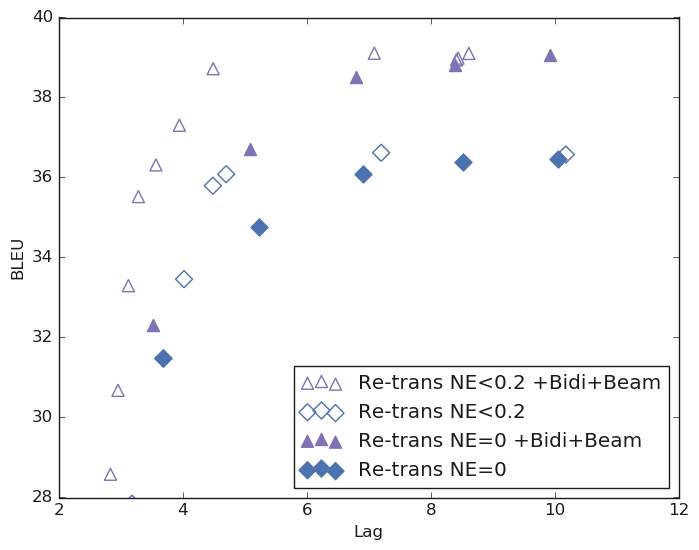}
    \caption{BLEU vs lag (\EDAL{}) curves for re-translation with improved models for DeEn (left) and EnFr (right) test sets. 
    }
    \label{fig:bleu_vs_eal_extend}
\end{figure*}
Motivated by the strong performance of re-translation with few revisions, we now evaluate it with no
revisions, by setting $\beta$ to 1, which guarantees NE $=0$.
Since $\beta$ is locked at 1, we can build a curve by varying $k$ from 2 to 10 in increments of 2.
In this setting, re-translation becomes equivalent to wait-$k$ inference without wait-$k$ training, which is studied as an ablation to wait-$k$ training by \newcite{ma-etal-2019-stacl}.\footnote{Re-translation with beam search and with $\beta=1$ is similar to wait-$k$ inference with speculative beam search~\cite{zheng-etal-2019-speculative}, due to effective look-ahead from implementing wait-$k$ with truncation. However, we only evaluate greedy search in this comparison, where their equivalence is exact.}
However, where they tested wait-$k$ inference on a system with full-sentence training, we do so for a system with proportional prefix training (\S~\ref{sec:PropAlign}).
As before, we compare to our streaming baselines, test only our \textit{Base} system, and include a no-prefix ablation corresponding to full-sentence training.

Results are shown in Figure ~\ref{fig:bleu_lag_sim_streaming}.
First, re-translation outperforms wait-$k$ training at almost all latency levels. This is startling, because each wait-$k$ training point is trained specifically for its $k$, while the re-translation points reflect a single training run, reconfigured for different latencies by adjusting $k$ at test time. We suspect that this improvement stems  from prefix-training introducing stochasticity to the amount of source context used to predict target words, making the model more robust.
Second, without prefix training, re-translation is consistently below wait-$k$ training, confirming earlier experiments by \newcite{ma-etal-2019-stacl} on the ineffectiveness of wait-$k$ inference without specialized training, and confirming our earlier observations on the surprising effectiveness of prefix training. 
Finally, we see that even without revisions, re-translation is very close to MILk, suggesting that this combination of prefix training and wait-$k$ inference is an extremely strong baseline, even for a 0-revision regime.

\subsection{Extendability of re-translation}

Re-translation's primary strengths lie in its ability to revise and its ability to apply to any MT system. 
With some effort, streaming systems can be fitted with enhancements such as bidirectional encoding~\cite{ma-etal-2019-stacl},\footnote{Any streaming model with a bidirectional encoder requires re-encoding for each source prefix, resulting in higher compute and memory costs.} beam search~\cite{zheng-etal-2019-speculative} and multihead attention~\cite{ma-etal-2019-montonic}.
Conversely, re-translation can wrap any auto-regressive NMT system and immediately benefit from its improvements.
Furthermore, re-translation's latency-quality trade-off can be manipulated without retraining the base system.
It is not the only solution to have these properties; most policies that are not trained jointly with NMT can make the same claims~\cite{Cho16,Gu2017,zheng-etal-2019-simpler}.
We conduct an experiment to demonstrate the value of this flexibility, by comparing our \textit{Base} system to the upgraded \textit{Bidi+Beam}.\footnote{We could just as easily upgrade to a different base architecture, such as the Transformer~\cite{vaswani2017attention}, which could potentially to lead to further improvements.} We carry out this test with few revisions (NE $< 0.2$) and without revisions (NE $=0$), projecting Pareto curves from dev to test where necessary.
The results are shown in Figure~\ref{fig:bleu_vs_eal_extend}.

Comparing the few-revision (NE $<0.2$) curves, we see large improvements, some more than 2 BLEU points, from using better models. Looking at the no-revision (NE $=0$) curves, we see that this configuration also benefits from modeling improvements, but for DeEn, the deltas are noticeably smaller than those of the few-revision curves.

\subsection{On computational complexity}
Re-translation is conceptually simple and easy to implement, but also incurs an increase in asymptotic time complexity.
If the base model can translate a sentence in time $O(x)$, then re-translation takes $O(nx)$ where $n$ is the number of times we request re-translation for that sentence. $n$ is capped at the length of the sentence,  as we never revise translations of earlier sentences in the transcript.\footnote{Though our experiments in this paper are sentence-level for the sake of continuity with earlier work on streaming translation, in practice, source sentence boundaries can be provided by speech-recognition endpointing or unspoken punctuation prediction~\cite{arivazhagan2019retranslation}.}

For many settings, this increase in complexity can be easily ignored.  
We are not concerned with the total time to translate a sentence, but instead with the latency between a new source word being uttered and its translation's appearance on the screen.
Modern accelerators can translate a complete sentence in the range of 100 milliseconds,\footnote{24.43 milliseconds per sentence is achievable on a NVIDIA TITAN RTX, according to MLPerf v0.5 Inference Closed NMT Stream. Retrieved from www.mlperf.org November 6th, 2019, entry Inf-0.5-27. MLPerf name and logo are trademarks. See www.mlperf.org for more information.} meaning that the time required to update the screen by translating an updated source prefix is small enough to be imperceptible.
As in all simultaneous systems, the largest source of latency is waiting for new source content to arrive.\footnote{We acknowledge the desirability of low computational complexity, both for reducing the cumulative burden alongside other sources of latency, such as network delays, and for reducing power consumption for mobile or green computing. Finding a lower-cost solution that preserves the flexibility of re-translation is a worthy goal for future research.}

\section{Conclusion}

We have presented the first comparison of re-translation and streaming strategies for simultaneous translation.
We have shown re-translation with low levels of erasure ($\mathrm{NE}<0.2$) to be as good or better than the state of the art in streaming translation. Also, re-translation easily embraces arbitrary improvements to NMT, which we have highlighted with large gains from an upgraded base model.

In our setting, re-translation with no erasure reduces to wait-$k$ inference, which we have shown to be much more effective than previously reported, so long as the underlying NMT system's training data has been augmented with prefix pairs.
Due to its simplicity and its effectiveness, we suggest re-translation as a strong baseline for future research on simultaneous translation.

\bibliography{acl2020}
\bibliographystyle{acl_natbib}

\end{document}